\documentclass[conference]{IEEEtran}
\IEEEoverridecommandlockouts

\usepackage{soul}
\usepackage{enumitem}
\usepackage{mathtools}
\usepackage{empheq}
\usepackage{graphicx}
\usepackage{epstopdf}
\usepackage{svg}
\usepackage[normalem]{ulem}
\usepackage{url}
\usepackage{amsmath}
\usepackage{breakurl}
\usepackage{hyperref}
\usepackage{subcaption}
\hypersetup{
    colorlinks=true,
    linkcolor=black,
    citecolor=black,
    urlcolor=black,
    pdfborder={0 0 0},
}
\graphicspath{{figures/}}
\usepackage{xspace}

\makeatletter
\def\abstract{\normalfont
    \if@twocolumn
      \@IEEEabskeysecsize\noindent\bfseries\textit{\abstractname}---\ignorespaces
    \else
      \bgroup\par\addvspace{0.5\baselineskip}\centering\vspace{-1.78ex}\@IEEEabskeysecsize\textbf{\abstractname}\par\addvspace{0.5\baselineskip}\egroup\quotation\@IEEEabskeysecsize
    \fi\bfseries\ignorespaces}
\def\endabstract{\relax\ifCLASSOPTIONconference\vspace{0ex}\else\vspace{1.34ex}\fi\par\if@twocolumn\else\endquotation\fi
    \normalfont\normalsize}
\def\IEEEkeywords{\normalfont
    \if@twocolumn
      \@IEEEabskeysecsize\vskip 0.5\baselineskip plus 0.25\baselineskip minus 0.25\baselineskip\noindent
      \bfseries\textit{\IEEEkeywordsname}---\ignorespaces
    \else
      \bgroup\par\addvspace{0.5\baselineskip}\centering\@IEEEabskeysecsize\textbf{\IEEEkeywordsname}\par\addvspace{0.5\baselineskip}\egroup\quotation\@IEEEabskeysecsize
    \fi\bfseries\ignorespaces}
\def\endIEEEkeywords{\relax\ifCLASSOPTIONtechnote\vspace{1.34ex}\else\vspace{0.67ex}\fi
    \par\if@twocolumn\else\endquotation\fi%
    \normalfont\normalsize}
\makeatother
\begin{document}
\title{Utility-Guided Orchestration for Cost-Efficient Tool-Augmented LLM Services}
\newcommand{\ustc}{University of Science and Technology of China, Hefei, Anhui, China}
\author{
  \IEEEauthorblockN{Boyan Liu\textsuperscript{1,4},
  Gongming Zhao\textsuperscript{1,4},
  Hongli Xu\textsuperscript{1,4},
  Huihui Tang\textsuperscript{2},
  Wenpeng Zhu\textsuperscript{3},
  Li Li\textsuperscript{3},
  Yuanpu Zhang\textsuperscript{3}
  }
  \IEEEauthorblockA{
    \textsuperscript{1} \ustc\\
    \textsuperscript{2} Guangxi Key Laboratory of Digital Infrastructure, Guangxi Zhuang Autonomous Region Information Center\\
    \textsuperscript{3} China Mobile (Suzhou) Software Technology Company Ltd., Suzhou 518048, China\\
    \textsuperscript{4} Suzhou Institute for Advanced Research, University of Science and Technology of China
  }
}

\maketitle
\vspace{-5mm}

\begin{abstract}
Tool-augmented large language model (LLM) services can solve complex tasks through retrieval and external tools, but current execution paradigms often trade adaptability for efficiency. Fixed workflows are predictable but rigid, while free-form reasoning loops such as ReAct may over-execute and issue redundant tool calls. We propose a lightweight utility-guided orchestration framework that formulates agent control as a cost-aware sequential decision problem over a compact action space: respond, retrieve, tool call, verify, and stop. An interpretable utility function balances expected gain, step-cost proxies, uncertainty, and redundancy. Experiments on multi-hop question answering show that the policy offers a controllable quality--cost trade-off and reduces token consumption by up to 10.6\% in the semantic-redundancy setting while preserving similar answer quality. The framework is intended as an inspectable control layer for practical LLM services rather than a universally dominant accuracy optimizer.
\end{abstract}

\begin{IEEEkeywords}
LLM Agents, Tool use, Orchestration policy, Cloud AI services
\end{IEEEkeywords}

\section{INTRODUCTION}

Large language models (LLMs) are evolving from standalone text
generators into interactive systems that reason, retrieve evidence,
call tools, and operate in external environments
\cite{yao2023react,schick2023toolformer,patil2023gorilla,
lewis2020rag,nakano2021webgpt,shen2023hugginggpt,park2023generative}.
Recent work shows that multi-step reasoning, planning, program use,
and reflection can improve complex task solving
\cite{karpas2022mrkl,gao2023pal,wang2023planandsolve,xu2023rewoo,
wei2022chain,wang2023selfconsistency,yao2023tree,chen2023program,
madaan2023selfrefine,shinn2023reflexion,zhou2024lats,wang2023voyager}.
Agent benchmarks such as WebShop, ALFWorld, and ToolQA further show
that evaluation must consider not only final answers but also the
efficiency and controllability of intermediate trajectories
\cite{yao2022webshop,shridhar2021alfworld,zhuang2023toolqa}. This is
especially important in cloud services, where token budgets, latency,
and adaptive computation are practical constraints
\cite{tay2022efficient,graves2016adaptive,elbayad2020depth,
schuster2022confident,xin2020deebert}.

Existing systems mainly follow two execution paradigms. Fixed workflows
predefine action sequences through prompts, roles, or pipelines, as in
MetaGPT, ChatDev, AutoGen, and DSPy
\cite{hong2024metagpt,qian2024chatdev,wu2023autogen,khattab2023dspy}.
They provide predictable costs but adapt poorly to query difficulty.
Free-form reasoning loops such as ReAct, Toolformer, Gorilla, and
ToolLLM let the model decide when to reason and invoke tools
\cite{yao2023react,schick2023toolformer,patil2023gorilla,
qin2023toolllm}. They are adaptive, but can produce long trajectories
and redundant retrieval or tool calls.

Although these paradigms have achieved encouraging results, they exhibit critical limitations in practical cloud deployments:

\begin{itemize}
    \item \textbf{Limited adaptability.} Fixed pipelines cannot adjust execution to the marginal utility of new evidence.
    \item \textbf{Over-execution.} Free-form agents may generate long trajectories and redundant tool calls, increasing token use and latency without proportional accuracy gains.
    \item \textbf{Implicit stopping.} Prompt-driven continuation and termination make the quality--cost trade-off difficult to inspect or regulate.
\end{itemize}

To decouple reasoning from execution control, this work introduces a
utility-guided orchestration framework for tool-augmented LLM services.
Agent execution is modeled as an explicit decision process over a compact
action space, with \texttt{stop} and \texttt{verify} treated as first-class
actions. A lightweight utility function combines gain, cost, uncertainty,
and redundancy signals, forming an interpretable interface between the LLM
and its environment.

The main contributions of this work are summarized as follows:

\begin{itemize}
    \item We propose a lightweight orchestration framework that shifts agent execution from implicit prompt-driven behavior to an \textbf{explicit utility-guided decision process}.
    \item We design a utility-based action selection mechanism that jointly optimizes for \textbf{estimated gain, cost proxies, uncertainty, and redundancy}, enabling precise regulation of stopping decisions.
    \item Controlled experiments demonstrate a competitive and inspectable quality--cost trade-off, including up to a \textbf{10.6\% reduction in token consumption} in the semantic-redundancy setting while preserving similar answer quality.
    \item We analyze cost definitions, redundancy control, and ablations, clarifying the \textbf{controllability and efficiency} of modern LLM agents.
\end{itemize}

\section{MOTIVATION}

Tool-augmented LLM agents often solve complex tasks
through iterative reasoning and external tool
invocations. Existing frameworks, however, can produce
unnecessarily long trajectories. Redundant reasoning steps
increase token consumption, latency, and prompt context growth,
which directly affects practical LLM service efficiency.

To illustrate this issue, consider a typical multi-hop
question from the HotpotQA dataset:

\textit{Question: Which university did the founder of
SpaceX attend?}

The answer requires identifying the founder of SpaceX
(Elon Musk) and then the university he attended
(University of Pennsylvania). Ideally, only a few retrieval
operations are needed. In practice, our HotpotQA analysis
shows that ReAct-style agents use 3.47 reasoning steps per
query on average, although many questions can be solved within
two retrieval operations. Extra steps often repeat related
entities, rephrase the same sub-question, or retrieve irrelevant
documents. In our experiments, semantic redundancy control
reduces token consumption by 10.6\% while maintaining similar
answer quality, indicating that a meaningful portion of cost
comes from avoidable intermediate actions.

These observations motivate the design of a
lightweight orchestration mechanism that explicitly
controls agent execution. Instead of relying solely
on fixed pipelines or unconstrained reasoning loops,
the proposed framework evaluates candidate actions
based on their potential utility and execution cost.
By explicitly balancing reasoning benefit and
computational overhead, the agent can avoid redundant
operations while maintaining effective multi-step
reasoning capability.

\section{DESIGN}
\begin{figure}[t]
    \centering
    \includegraphics[width=\columnwidth]{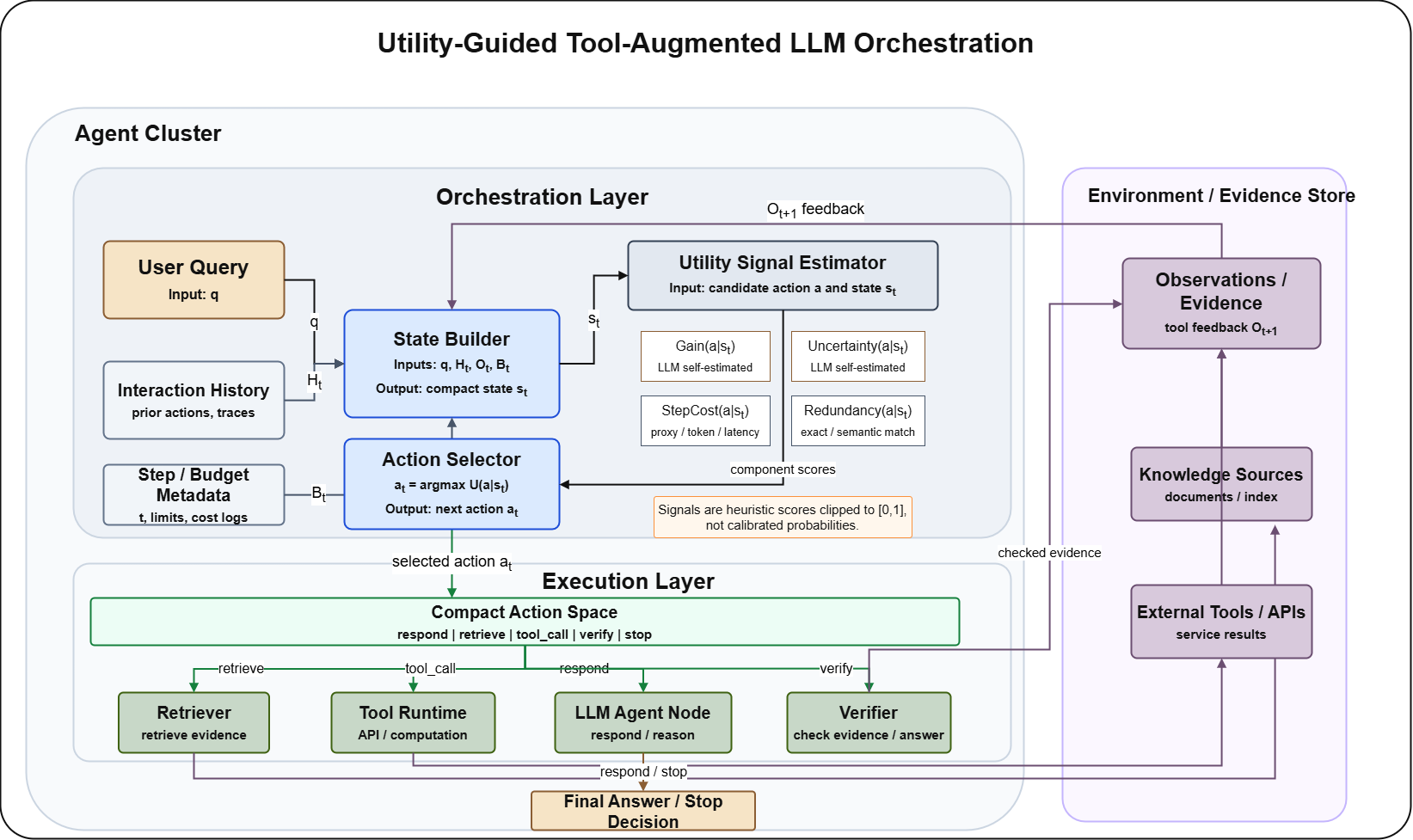}
    \caption{Utility-Guided Agent System Overview}
    \label{fig:framework}
\end{figure}

Figure~\ref{fig:framework} illustrates the proposed orchestration framework. The goal is not to replace the reasoning capability of the underlying LLM, but to add an explicit control layer that decides when the agent should continue reasoning, which operation should be executed next, and when execution should terminate. Compared with fixed workflows, this design adapts to query difficulty and intermediate evidence; compared with unconstrained reasoning loops, it exposes cost and redundancy as first-class control signals.

\subsection{Problem Formulation}
We model tool-using agent orchestration as a sequential decision problem. Given a user query $q$, the agent observes state $s_t$ at step $t$ and selects the next action $a_t$. The action space is intentionally compact:
\begin{equation}
A = \{respond, retrieve, tool\_call, verify, stop\}.
\end{equation}
\texttt{respond} generates an answer from the current state; \texttt{retrieve} obtains external evidence; \texttt{tool\_call} invokes an external API or computation tool; \texttt{verify} checks the current answer or evidence; and \texttt{stop} terminates execution. This space covers the major control decisions in practical tool-augmented agents while remaining small enough to analyze.

The controller repeats action selection until \texttt{stop} is chosen or a predefined execution limit is reached. The objective is to obtain a sufficiently accurate trajectory without unnecessary reasoning length, redundant tool usage, or excessive token cost. Thus, agent execution is treated as a quality--cost trade-off rather than a pure reasoning problem.

The compact action space separates semantic task solving from systems-level orchestration. The LLM may still generate rich reasoning traces or tool arguments internally, while the controller decides only whether the next operation is answering, retrieving, invoking a tool, verifying, or stopping. Modeling \texttt{verify} and \texttt{stop} explicitly lets the controller compare them with retrieval and response generation under the same utility scale.

\subsection{Agent State Representation}
At step $t$, the state $s_t$ contains four categories of information: the original query and current working context; prior actions and intermediate reasoning traces; external observations returned by retrieval or tools; and execution-status signals such as step count and budget metadata. This explicit state lets the controller distinguish, for example, whether evidence is still missing, whether recent retrieval has already produced enough information, and whether the trajectory is becoming overly long.

The state is not intended to be a fully learned world model. It is a structured control interface between the LLM and the environment, designed to make continuation and stopping decisions inspectable.

The state representation is deliberately compact: it stores what the user asked, what has already been tried, what evidence has been observed, and how much budget has been consumed. This avoids turning the controller itself into a new source of prompt growth while retaining the signals most relevant to redundant execution and premature stopping.

\subsection{Utility-Guided Action Selection}
Each candidate action $a \in \mathcal{A}$ receives a utility score:
\begin{equation}
\begin{split}
U(a \mid s_t) = & \mathrm{Gain}(a \mid s_t) \\
& - \lambda_1 \mathrm{StepCost}(a \mid s_t) \\
& - \lambda_2 \mathrm{Uncertainty}(a \mid s_t) \\
& - \lambda_3 \mathrm{Redundancy}(a \mid s_t).
\end{split}
\label{eq:utility}
\end{equation}
The next action is selected by
\begin{equation}
a_t^\star = \arg\max_{a \in \mathcal{A}} U(a \mid s_t).
\label{eq:action_selection}
\end{equation}
We do not claim that Eq.~(\ref{eq:utility}) is an optimal reinforcement-learning policy or that its terms are calibrated probabilities. Instead, the utility function is a lightweight, interpretable scoring mechanism that turns implicit prompt-driven behavior into an explicit decision process. A useful action should have high expected benefit and low expected cost; repeated or low-novelty actions should be penalized.

The explicit utility also separates reasoning from execution control. The LLM remains responsible for generating answers, interpreting evidence, and interacting with tools, while the controller decides whether another operation is worth performing. This separation makes it possible to inspect which term pushes a trajectory to continue, stop, retrieve, or verify. It also allows alternative cost definitions to be substituted without redesigning the whole agent.

Operationally, the controller evaluates all candidate actions with the same score, so a selected action can be traced to a small set of component values and weights. The weights $\lambda_1$, $\lambda_2$, and $\lambda_3$ tune the relative strength of cost, uncertainty, and redundancy penalties for different deployment requirements.

This design is intentionally heuristic rather than fully learned. A learned policy could in principle optimize these decisions from trajectory rewards, but it would require a larger training set, stable reward definitions, and retraining when tool prices or service-level objectives change. The utility form instead exposes a small number of knobs that can be adjusted by operators. For example, a low-latency service can increase the cost weight, while an analysis-oriented service can relax the cost penalty and increase uncertainty control.

\subsection{Utility Components}
\paragraph{Estimated Gain.}
$\mathrm{Gain}(a \mid s_t)$ measures the self-estimated marginal value of executing action $a$. When evidence is incomplete, \texttt{retrieve} or \texttt{tool\_call} may have high gain; when evidence appears sufficient, \texttt{respond} or \texttt{stop} may be favored. This heuristic signal preserves the ability to continue on genuinely difficult queries.

\paragraph{Step-Cost.}
$\mathrm{StepCost}(a \mid s_t)$ is a normalized proxy for the cost of taking another step. It discourages unnecessary trajectory expansion while keeping the controller cheap to compute. It is not an exact substitute for token usage or wall-clock latency; Section IV compares it with token- and latency-aware variants.

We use a proxy because exact runtime and token cost may only be known after an action finishes. A controller that needs those exact measurements before every decision would itself introduce overhead. The proxy therefore acts as a lightweight regularizer that biases the agent away from unnecessary continuation, while the empirical evaluation checks whether this bias is aligned with observed cost.

\paragraph{Uncertainty.}
$\mathrm{Uncertainty}(a \mid s_t)$ reflects the agent's self-estimated confidence in the sufficiency of current evidence. Higher uncertainty penalizes premature termination, while lower uncertainty favors responding or stopping. This prevents a cost-aware controller from becoming overly aggressive.

\paragraph{Redundancy.}
$\mathrm{Redundancy}(a \mid s_t)$ penalizes repeated or semantically similar actions. It targets a common failure mode of free-form agents: issuing similar retrieval or tool calls after useful evidence has already been obtained. The term reduces token consumption and makes trajectories more compact and interpretable.

Redundancy can be measured exactly, by repeated actions or retrieval strings, or semantically, by comparing the meaning of the current candidate action with previous actions or observations. Exact matching is cheap; semantic matching catches paraphrased repetitions at higher overhead.

Together, these four terms are not meant to form a universal decomposition of agent intelligence. They provide a practical control surface for studying the quality--cost trade-off of tool-using LLM systems. This framing is especially important for deployment settings where final answer quality must be considered together with latency, token budget, and trajectory stability.

The same formulation can also represent heterogeneous tool costs. Although the present implementation uses a single retrieval interface, production agents may combine local retrieval, remote search, database access, code execution, function calling, and additional model invocations. These operations have different latency, price, failure modes, and retry behavior. A utility-guided controller can encode such differences through action-specific cost terms or tool-specific redundancy penalties. For instance, a cheap local retrieval action can be allowed more often, while an expensive remote API call can require higher expected gain or higher uncertainty before being selected. This keeps the controller portable while making cost assumptions explicit.

\subsection{Implementation Details}
The controller is implemented as a wrapper around a standard tool-augmented LLM loop. At each step, it serializes the user query, recent actions, retrieved observations, and step metadata into a compact controller prompt. The LLM estimates the heuristic signals, which are clipped to $[0,1]$ before scoring so that no single unbounded self-report dominates the decision. Candidate actions are scored under the same utility template: evidence-seeking actions such as \texttt{retrieve} or \texttt{tool\_call} receive gain from expected new information, while \texttt{respond} and \texttt{stop} receive gain from estimated state sufficiency. After selection, the execution module performs the action and appends the observation to the state. This keeps the controller separate from the model and retriever, making it possible to replace either component or add tools without changing the surrounding evaluation harness.

\subsection{Execution Loop and Termination}
Starting from the user query, the system constructs an initial state, scores all candidate actions, selects the highest-utility action, executes it, and updates the state with the resulting observation. Termination occurs when \texttt{stop} is selected, a step budget is exhausted, or a fallback condition is triggered. Because stopping is governed by the same utility logic as retrieval, tool use, and verification, the final trajectory is bounded, inspectable, and suitable for cost-sensitive deployment.

Each selected action can be logged with its component scores, making post-hoc analysis more direct. Operators can inspect whether an expensive trajectory came from high uncertainty, repeated retrieval with weak redundancy penalties, or an insufficient cost weight, then update the corresponding utility component instead of rewriting the entire prompt or agent loop.

\section{Evaluation}
This section evaluates the proposed framework from four perspectives:
overall quality--cost trade-off, reasoning depth, workflow fairness, and
utility-component behavior.

\subsection{Experimental Setup}
We evaluate all methods on a fixed sample of 200 HotpotQA development examples, using the same base model, local BM25 retriever, and sampled question set throughout. Experiments run on Ubuntu 22.04.5 with dual Intel Xeon Gold 6330 CPUs, 1\,TB memory, and 8 NVIDIA A800-SXM4-40 GB GPUs. The language model is served through a vLLM-based OpenAI-compatible endpoint, and retrieval uses a local BM25 index over the HotpotQA corpus. We report F1, token consumption, wall-clock time, and $F1 / \text{tokens}$.

The compared methods include \texttt{direct} answering, fixed \texttt{workflow} baselines, a \texttt{threshold} controller, \texttt{ReAct}, and our utility-guided \texttt{policy}. For the policy, \texttt{expected\_gain} and \texttt{uncertainty} are clipped LLM self-estimates rather than calibrated probabilities. The default cost term is \texttt{step\_cost}; token- and latency-based variants are controlled comparisons. The implementation and scripts are open source in our \href{https://github.com/kagami-kasumi/Utility-Guided-Agent-Orchestration-for-Efficient-LLM-Tool-Use}{GitHub repository}.

\subsection{Main Results}
\begin{figure*}[t]
\centering
\begin{subfigure}{0.32\textwidth}
\centering
\includegraphics[width=\linewidth]{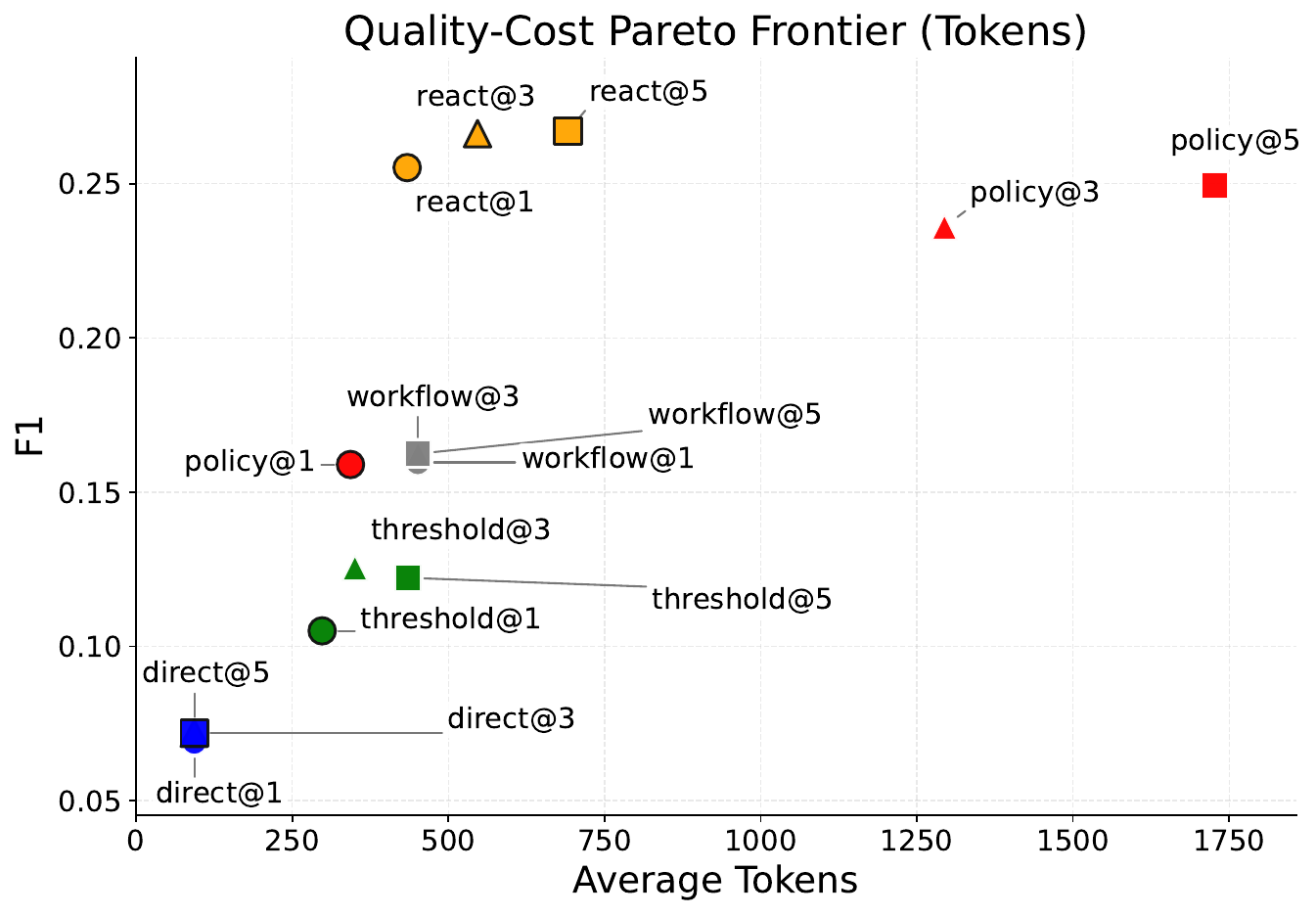}
\caption{Token-based quality--cost trade-off}
\end{subfigure}
\hfill
\begin{subfigure}{0.32\textwidth}
\centering
\includegraphics[width=\linewidth]{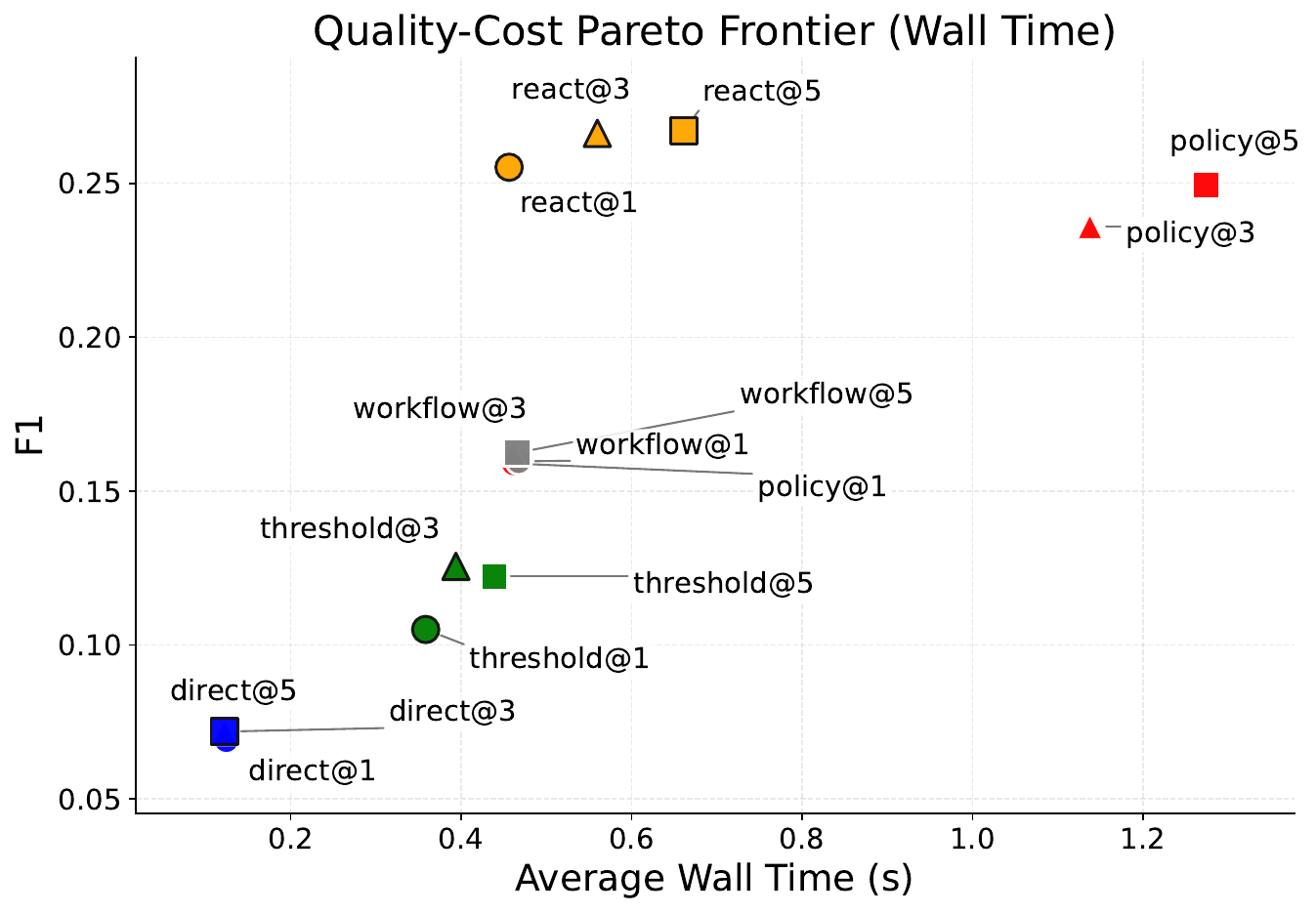}
\caption{Latency-based quality--cost trade-off}
\end{subfigure}
\hfill
\begin{subfigure}{0.32\textwidth}
\centering
\includegraphics[width=\linewidth]{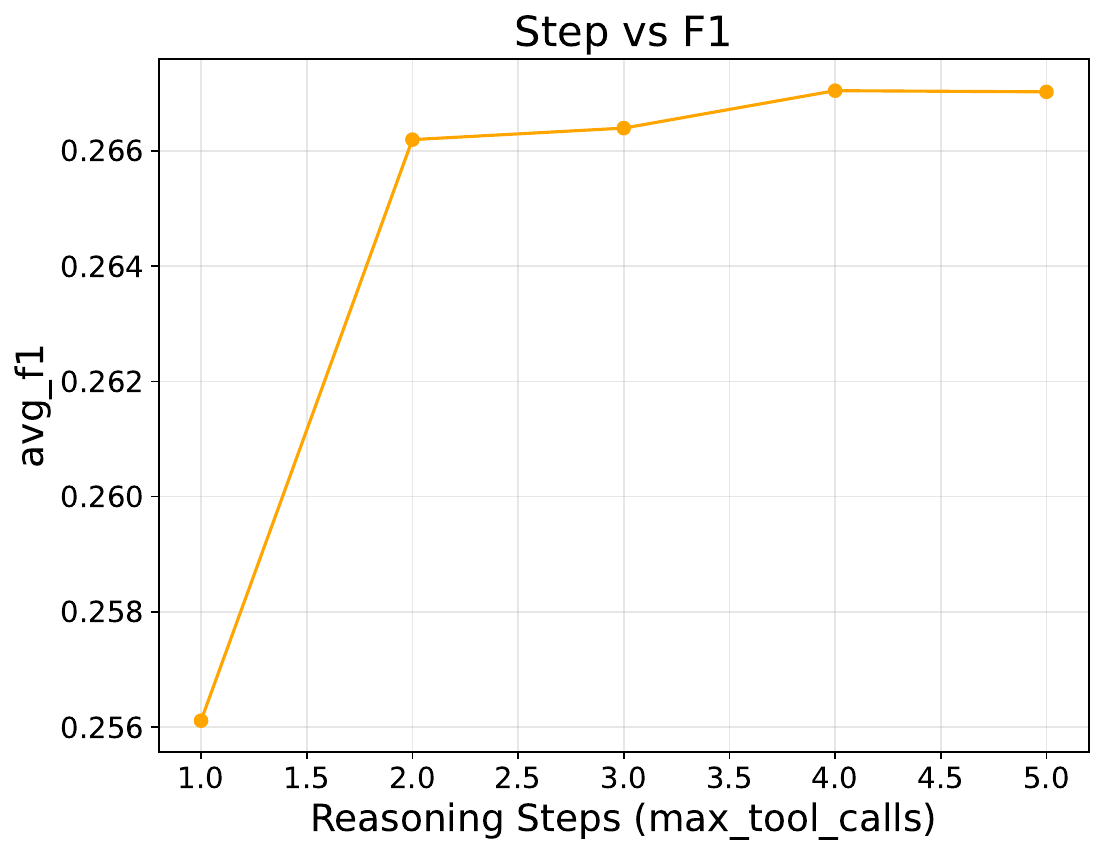}
\caption{Reasoning depth vs answer quality}
\end{subfigure}
\caption{Quality--cost trade-offs and reasoning-depth analysis.}
\label{fig:main_results}
\end{figure*}

Figures~\ref{fig:main_results}(a) and~\ref{fig:main_results}(b) summarize the global quality--cost trade-off from token and latency perspectives, while Table~\ref{tab:main_results} reports exact numbers.

\begin{table}[t]
\centering
\small
\caption{MAIN RESULTS}
\resizebox{\columnwidth}{!}{
\begin{tabular}{lrrrr}
\hline
Method & F1 & Tokens & Wall Time & Efficiency \\
\hline
direct & 0.0719 & 93.0 & 0.122 & 0.000772 \\
workflow (minimal) & 0.1625 & 451.2 & 0.461 & 0.00036 \\
workflow-search-twice & 0.1698 & 514.1 & 0.902 & 0.00033 \\
workflow-search-verify & 0.063 & 1041.2 & 1.617 & 6.1e-05 \\
threshold & 0.1255 & 350.3 & 0.394 & 0.000358 \\
react & 0.2662 & 546.6 & 0.56 & 0.000487 \\
policy (step-cost) & 0.236 & 1294.2 & 1.138 & 0.000182 \\
\hline
\end{tabular}
}
\end{table}

\texttt{ReAct} achieves the strongest F1 ($0.2662$), confirming the value of flexible multi-step reasoning. Our \texttt{policy(step\_cost)} reaches $F1=0.2360$ and exposes continuation, stopping, and redundant retrieval through explicit utility terms. The result therefore supports the policy as an interpretable quality--cost controller, not as a universal accuracy replacement for free-form reasoning.

\subsection{Reasoning Depth Analysis}
Figure~\ref{fig:main_results}(c) shows that additional reasoning steps initially improve answer quality, but with diminishing returns as token usage and latency rise. This supports treating orchestration as a decision over whether another step is worth its marginal cost.

\begin{table}[t]
\centering
\small
\caption{COST DEFINITION COMPARISON}
\resizebox{\columnwidth}{!}{
\begin{tabular}{lrrrr}
\hline
Method & F1 & Tokens & Wall Time & Efficiency \\
\hline
workflow (minimal) & 0.1625 & 451.2 & 0.461 & 0.00036 \\
react & 0.2662 & 546.6 & 0.56 & 0.000487 \\
threshold & 0.1255 & 350.3 & 0.394 & 0.000358 \\
policy (step-cost) & 0.236 & 1294.2 & 1.138 & 0.000182 \\
policy (token-cost) & 0.2562 & 1308.6 & 1.215 & 0.000196 \\
policy (latency-cost) & 0.2447 & 1272.9 & 1.152 & 0.000192 \\
\hline
\end{tabular}
}
\end{table}

The cost-definition comparison clarifies the role of \texttt{step\_cost}. Token- and latency-aware variants reach higher F1 ($0.2562$ and $0.2447$), while all variants remain in a similar cost regime. Thus, the step proxy is a deployment-friendly default that can be replaced by precise instrumentation without changing the decision framework.

\subsection{Workflow Fairness Analysis}
\begin{table}[t]
\centering
\small
\caption{WORKFLOW FAIRNESS}
\resizebox{\columnwidth}{!}{
\begin{tabular}{lrrrrr}
\hline
Method & F1 & Tokens & Wall Time & Efficiency & Tool Calls \\
\hline
workflow (minimal) & 0.1625 & 451.2 & 0.461 & 0.00036 & 1.0 \\
workflow-search-twice & 0.1698 & 514.1 & 0.902 & 0.00033 & 2.0 \\
workflow-search-verify & 0.063 & 1041.2 & 1.617 & 6.1e-05 & 2.0 \\
\hline
\end{tabular}
}
\end{table}

To address the concern that the workflow baseline is too weak, we add \texttt{workflow-search-twice} and \texttt{workflow-search-verify}. The former slightly improves F1 ($0.1698$ vs.\ $0.1625$) but nearly doubles wall time, while the latter performs worse and is more expensive. Fixed extra retrieval or verification can help some difficult cases, but it also wastes cost on easy cases, motivating state-dependent orchestration.

This comparison is included to make the evaluation fairer to fixed pipelines. A minimal workflow alone could make the proposed controller look stronger simply because the baseline is under-specified. The two augmented workflows test whether adding more fixed actions is enough to close the gap. The results suggest that the issue is not merely the number of steps, but the inability to choose steps conditionally. Some questions benefit from extra retrieval or verification, while others are already answerable after the first evidence item. A utility controller can express this distinction at inference time.

\subsection{Redundancy Analysis}
\begin{table}[t]
\centering
\small
\caption{EXACT vs.\ SEMANTIC REDUNDANCY CONTROL}
\resizebox{\columnwidth}{!}{%
\begin{tabular}{lrrrrr}
\hline
Method & F1 & Tokens & Wall Time & Tool Calls & Redundant Tool Calls \\
\hline
policy (step\_cost) & 0.2360 & 1294.2 & 1.138 & 1.56 & 0.44 \\
policy (semantic redundancy) & 0.2370 & 1156.6 & 1.346 & 1.40 & 0.43 \\
\hline
\end{tabular}
}
\end{table}

Semantic redundancy control preserves answer quality ($0.2370$ vs.\ $0.2360$ F1) while reducing tokens from $1294.2$ to $1156.6$ and tool calls from $1.56$ to $1.40$. Wall time increases from $1.138$ to $1.346$ because semantic comparison adds local overhead, so the current benefit is token and trajectory compactness rather than runtime acceleration. The unchanged F1 suggests that many removed operations are low-value repetitions.

This result is also useful for interpreting the deployment value of redundancy control. In our local retrieval setting, a repeated retrieval call is not very expensive, so reducing tool calls does not automatically reduce wall-clock time. In hosted LLM services, however, repeated actions often consume paid tokens, expand the prompt context, and may call remote APIs whose latency is outside the model server. Under such conditions, trajectory compactness can have operational value even when the local benchmark shows only token savings. The semantic variant should therefore be read as a conservative demonstration: it removes repeated behavior without a measurable quality drop, while leaving lower-overhead redundancy detection as an engineering optimization.

\subsection{Heuristic Signals and Ablations}
\begin{figure}[t]
    \centering
    \includegraphics[width=0.80\linewidth]{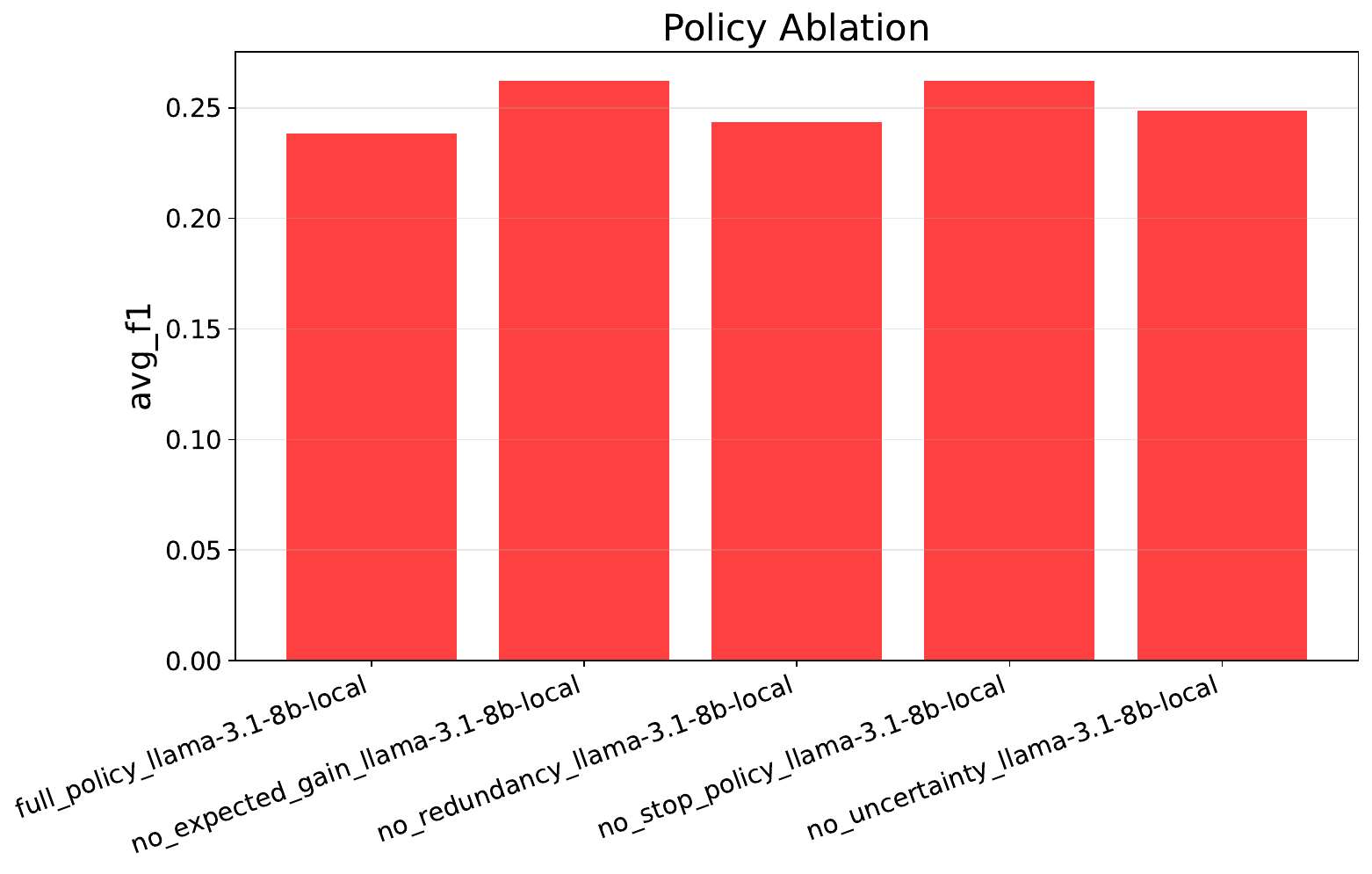}
    \caption{Policy ablation results. Removing gain-, uncertainty-, redundancy-, or stop-related control generally pushes the agent toward a less favorable quality--cost regime.}
    \label{fig:policy-ablation}
\end{figure}

\begin{table}[t]
\centering
\small
\caption{POLICY ABLATION}
\resizebox{\columnwidth}{!}{
\begin{tabular}{lrrrr}
\hline
Method & F1 & Tokens & Wall Time & Efficiency \\
\hline
full policy & 0.2383 & 1273.3 & 1.113 & 0.000187 \\
-expected-gain & 0.2621 & 2716.6 & 1.897 & 9.6e-05 \\
-uncertainty & 0.2487 & 1669.7 & 1.378 & 0.000149 \\
-redundancy & 0.2435 & 1792.7 & 1.403 & 0.000136 \\
-stop & 0.2621 & 2716.6 & 1.892 & 9.6e-05 \\
\hline
\end{tabular}
}
\end{table}

Expected gain is behaviorally meaningful: continue-rate is near zero in the lowest gain bucket and rises sharply in mid/high-gain ranges. Its Pearson correlation with final F1 is $0.1479$, while uncertainty is noisier ($0.0131$), so these values should be treated as control heuristics rather than calibrated probabilities. Ablations reinforce this interpretation. Removing gain, uncertainty, redundancy, or stop-related control usually increases cost more than quality; removing gain or stop raises average tokens above $2700$ while reaching only $F1=0.2621$. Overall, the evidence supports controllable orchestration in a controlled retrieval setting, with broader multi-tool validation left for future work.

Figure~\ref{fig:policy-ablation} provides the same pattern visually. The full policy is not the highest-F1 configuration in absolute terms, but it avoids the large token increase caused by removing gain or stop control. This matters for service deployment because operators usually need a stable operating point rather than an unconstrained search for marginal accuracy. The ablation therefore supports the central claim: explicit orchestration makes quality--cost behavior tunable and diagnosable, even when the underlying heuristic signals remain imperfect.

The evaluation therefore addresses two practical concerns: stronger fixed workflows show the cost of unconditional extra actions, and the semantic redundancy experiment separates token savings from latency reduction. Together, these results position the method as a controllable service mechanism rather than a claim of absolute dominance.

\section{Conclusion}
This paper presents a lightweight utility-guided orchestration framework for tool-augmented LLM agents. By scoring a compact action space with gain, cost, uncertainty, and redundancy signals, the framework turns continuation and stopping into explicit decisions. The resulting controller does not universally outperform strong free-form baselines, but it provides a competitive and inspectable mechanism for managing quality--cost trade-offs.

The utility components currently rely on self-estimated heuristic signals rather than calibrated probabilistic estimators, and the evaluation focuses on HotpotQA-style multi-hop question answering with a single retrieval tool. Future work will study learned utility estimation, lower-overhead redundancy detection, memory-aware orchestration, heterogeneous tool costs, and broader multi-tool tasks such as code execution, web automation, and API composition.

\section*{Acknowledgment}
The corresponding author of this paper is GongmingZhao. This work is supported by the National Natural Sci-ence Foundation of China (NSFC) under Grant 62372426, the Youth Innovation Promotion Association of the ChineseAcademy of Science under Grant 2023481, Basic ResearchProgram of Jiangsu under Grant BK20250124, the Funda-mental Research Funds for the Central Universities underGrant WK2150250043, and the Open Project Program of theGuangxi Key Laboratory of Digital Infrastructure (Grant No.GXDIOP2024003).
\bibliographystyle{IEEEtran}
\bibliography{main}
\end{document}